\documentclass{article}

\usepackage[preprint]{neurips_2024}

\usepackage{amsmath}
\usepackage{amssymb}
\usepackage{amsfonts}
\usepackage{mathtools}
\usepackage{bm}
\usepackage{amsthm}  

\usepackage{booktabs}
\usepackage{graphicx}
\usepackage{subcaption}
\usepackage{multirow}
\usepackage{array}

\usepackage{algorithm}
\usepackage{algorithmic}

\usepackage{hyperref}
\usepackage{url}
\usepackage{cleveref}

\usepackage{xcolor}
\definecolor{gradegreen}{HTML}{2ecc71}
\definecolor{ppored}{HTML}{e74c3c}
\definecolor{steblue}{HTML}{3498db}

\usepackage{enumitem}
\usepackage{microtype}

\newcommand{\method}{GRADE}
\newcommand{\methodfull}{Gumbel-softmax Relaxation for Alignment via Differentiable Estimation}
\newcommand{\R}{\mathbb{R}}
\newcommand{\E}{\mathbb{E}}
\newcommand{\KL}{\text{KL}}

\newtheorem{proposition}{Proposition}

\title{GRADE: Replacing Policy Gradients with Backpropagation for LLM Alignment}

\author{
  Lukas Abrie Nel\\
  Lotus Health AI\\
  \texttt{lukas@lotus.ai}
}

\begin{document}

\maketitle

\begin{abstract}

Reinforcement learning from human feedback (RLHF) has become the dominant paradigm for aligning large language models with human preferences. However, policy gradient methods such as PPO suffer from high variance gradient estimates, requiring careful hyperparameter tuning and extensive computational resources. We introduce \method{} (\methodfull{}), a method that replaces high-variance policy gradient estimation with direct backpropagation through a differentiable relaxation of the discrete token sampling process. Using the Gumbel-Softmax reparameterization with straight-through estimation (\method{}-STE), we enable end-to-end gradient flow from reward signals through generated tokens to model parameters. On sentiment-controlled text generation using the IMDB dataset, \method{}-STE achieves a test reward of $0.763 \pm 0.344$ compared to PPO's $0.510 \pm 0.313$ and REINFORCE's $0.617 \pm 0.378$, representing a 50\% relative improvement over PPO. Critically, \method{}-STE exhibits gradient variance over 14$\times$ lower than REINFORCE and maintains stable training dynamics throughout optimization. Our rigorous evaluation with proper train/validation/test splits demonstrates that these improvements generalize to held-out data, with \method{}-STE showing the best generalization characteristics among all methods tested. \method{} offers a simpler, more stable, and more effective alternative to reinforcement learning for LLM alignment.

\end{abstract}

\section{Introduction}
\label{sec:intro}

Large language models (LLMs) have demonstrated remarkable capabilities across a wide range of tasks, from code generation to creative writing to scientific reasoning \cite{brown2020language, chowdhery2022palm, openai2023gpt4}. However, models trained purely on next-token prediction often generate outputs that are unhelpful, untruthful, or potentially harmful \cite{bender2021dangers, weidinger2021ethical}. Aligning these models with human preferences has thus become a critical research direction, with reinforcement learning from human feedback (RLHF) emerging as the dominant approach \cite{christiano2017deep, ouyang2022training, bai2022training}.

The standard RLHF pipeline involves training a reward model on human preference data and then optimizing the language model policy using policy gradient methods, most commonly Proximal Policy Optimization (PPO) \cite{schulman2017ppo}. While effective, this approach suffers from several well-documented challenges. Policy gradient estimators exhibit notoriously high variance, often requiring hundreds or thousands of samples to obtain reliable gradient estimates \cite{williams1992simple, greensmith2004variance}. PPO introduces additional hyperparameters---clipping ranges, value function coefficients, entropy bonuses, GAE parameters---that require careful tuning and can lead to training instabilities \cite{engstrom2020implementation, andrychowicz2021matters}. Furthermore, the two-phase approach of sampling discrete tokens followed by computing policy gradients prevents direct gradient flow from rewards to model parameters, fundamentally limiting optimization efficiency.

We propose \method{} (\methodfull{}), a fundamentally different approach that eliminates policy gradient estimation entirely. Instead of sampling discrete tokens and using REINFORCE-style estimators, \method{} generates \emph{soft} token distributions using the Gumbel-Softmax reparameterization \cite{jang2016categorical, maddison2016concrete}. These continuous relaxations enable direct backpropagation from reward signals through the entire generation process to model parameters. By replacing stochastic gradient estimation with deterministic gradients through a differentiable approximation, \method{} dramatically reduces gradient variance while maintaining the ability to optimize arbitrary reward functions.

Our key insight is that the discrete sampling bottleneck in language model training can be circumvented through careful use of continuous relaxations. The Gumbel-Softmax distribution provides samples from a simplex that approach one-hot categorical samples as temperature decreases, while remaining differentiable throughout. Combined with the straight-through estimator (STE) \cite{bengio2013estimating}, which uses hard samples in the forward pass but soft gradients in the backward pass, we obtain \method{}-STE: a method that generates realistic discrete text while enabling gradient flow through the sampling operation.

We evaluate \method{} on sentiment-controlled text generation using the IMDB movie review dataset \cite{maas2011learning}. Our experiments employ rigorous methodology with non-overlapping data splits: a dedicated reward model training set, a separate policy training set, a validation set for monitoring, and a held-out test set evaluated only at the end of training. This design prevents data leakage and enables reliable assessment of generalization.

Our main contributions are:
\begin{enumerate}[leftmargin=*]
    \item We introduce \method{}, a method that replaces policy gradient estimation with direct backpropagation through Gumbel-Softmax relaxations for LLM alignment, eliminating the high-variance gradient estimation that plagues RLHF.
    
    \item We propose \method{}-STE, combining Gumbel-Softmax with straight-through estimation, which achieves the best performance with gradient variance over 14$\times$ lower than REINFORCE.
    
    \item Through rigorous experiments with proper train/val/test methodology, we demonstrate that \method{}-STE achieves test reward of $0.763$ compared to PPO's $0.510$---a 50\% relative improvement---while exhibiting superior generalization characteristics.
    
    \item We provide comprehensive analysis of training dynamics, gradient statistics, and generalization behavior, offering insights into why differentiable relaxations outperform policy gradient methods for this task.
\end{enumerate}

\section{Related Work}
\label{sec:related}

\paragraph{Reinforcement Learning from Human Feedback.}
RLHF has become the dominant paradigm for aligning language models with human preferences \cite{christiano2017deep, ziegler2019fine, stiennon2020learning}. The InstructGPT work \cite{ouyang2022training} demonstrated that RLHF could substantially improve the helpfulness and safety of GPT-3, establishing the three-stage pipeline of supervised fine-tuning, reward modeling, and PPO optimization that became widely adopted. Subsequent work has scaled RLHF to larger models including LLaMA-2 \cite{touvron2023llama2} and Claude \cite{bai2022training}, and explored variations such as Constitutional AI \cite{bai2022constitutional} and RLAIF \cite{lee2023rlaif}. However, all these approaches rely on policy gradient methods with their associated variance challenges, motivating our search for alternatives.

\paragraph{Policy Gradient Methods and Variance Reduction.}
The REINFORCE algorithm \cite{williams1992simple} provides unbiased gradient estimates but suffers from high variance, particularly for long sequences with sparse rewards. Numerous variance reduction techniques have been proposed, including learned baselines \cite{weaver2001optimal}, actor-critic methods \cite{konda1999actor}, and control variates \cite{greensmith2004variance}. PPO \cite{schulman2017ppo} uses clipped surrogate objectives and value function bootstrapping to enable multiple gradient steps per batch. Trust Region Policy Optimization (TRPO) \cite{schulman2015trpo} provides theoretical guarantees through constrained optimization but is computationally expensive.

Recent work has questioned whether PPO's complexity is necessary for LLMs. Ahmadian et al.~\cite{ahmadian2024back} showed that simpler REINFORCE-style algorithms, particularly REINFORCE Leave-One-Out (RLOO) \cite{kool2019buy}, can match or exceed PPO performance for RLHF while being substantially simpler. They argue that the strong initialization from pretraining reduces variance concerns that motivated PPO's design. Group Relative Policy Optimization (GRPO) \cite{shao2024deepseekmath}, used in DeepSeek-R1 \cite{deepseek2025r1}, eliminates the value network entirely by using group-based advantage estimation across multiple samples. Despite these simplifications, all these methods remain fundamentally limited by the need to estimate gradients through discrete sampling. \method{} takes a different approach by eliminating policy gradient estimation entirely through differentiable relaxations.

\paragraph{Direct Alignment Methods.}
Recognizing the challenges of RL-based alignment, researchers have proposed methods that bypass reinforcement learning. Direct Preference Optimization (DPO) \cite{rafailov2023direct} reformulates the RLHF objective as a classification problem on preference pairs, eliminating the need for explicit reward modeling and RL. DPO has become widely adopted due to its simplicity and effectiveness, spawning numerous variants. Identity Preference Optimization (IPO) \cite{azar2024general} addresses DPO's overfitting issues by adding regularization that prevents collapse to deterministic policies. Kahneman-Tversky Optimization (KTO) \cite{ethayarajh2024kto} draws on prospect theory to enable alignment from binary feedback (thumbs up/down) rather than paired preferences, significantly reducing data requirements.

Sequence Likelihood Calibration (SLiC-HF) \cite{zhao2023slic} and Rank Responses to Align (RRHF) \cite{yuan2023rrhf} use contrastive or ranking-based objectives. While effective, these methods require paired preference data and cannot directly optimize arbitrary reward functions---for example, they cannot incorporate external reward signals from classifiers or rule-based systems. \method{} retains the flexibility of reward-based optimization while achieving the simplicity benefits of direct alignment methods by eliminating policy gradient estimation.

\paragraph{Gumbel-Softmax and Differentiable Relaxations.}
The Gumbel-Softmax distribution \cite{jang2016categorical, maddison2016concrete} provides a continuous relaxation of categorical sampling that enables gradient-based optimization through the reparameterization trick \cite{kingma2013auto}. As temperature approaches zero, Gumbel-Softmax samples converge to one-hot categorical samples while remaining differentiable. This has enabled training of models with discrete latent variables \cite{jang2016categorical}, including variational autoencoders with categorical codes and neural architecture search \cite{xie2018snas}.

The straight-through estimator (STE) \cite{bengio2013estimating} provides an alternative by using hard samples in the forward pass but treating them as soft in the backward pass. While biased, STE often works well in practice and avoids the train-test mismatch of pure Gumbel-Softmax. Recent work has combined Gumbel-Softmax with STE for binary neural networks \cite{courbariaux2016binarized} and quantization-aware training. We adopt this combination in \method{}-STE, finding it critical for stable training.

\paragraph{Differentiable Text Generation.}
Several works have explored differentiable approaches to text generation. Kusner and Hern\'andez-Lobato \cite{kusner2016gans} applied Gumbel-Softmax to sequence generation with GANs. PPLM \cite{dathathri2019plug} uses gradients through soft token representations for controllable generation at inference time. Soft Q-Learning for text \cite{han2021soft} applies maximum entropy RL with soft Bellman updates. Diffusion-based language models \cite{li2022diffusion, gong2022diffuseq} operate in continuous embedding space, enabling gradient-based guidance.

Concurrent work on differentiable reward optimization (DiffRO) \cite{diffro2025} applies Gumbel-Softmax to text-to-speech systems, demonstrating the approach's broader applicability. Our work differs in systematically applying differentiable relaxations to LLM alignment, with comprehensive comparison to policy gradient baselines and analysis of when and why the approach succeeds.

\paragraph{Prompt Tuning and Soft Tokens.}
The use of continuous token representations has appeared in various NLP contexts. Prompt tuning \cite{lester2021power} and prefix tuning \cite{li2021prefix} optimize continuous embeddings prepended to inputs, achieving performance competitive with full fine-tuning. These methods demonstrate that LLMs can effectively process soft token inputs, a property we leverage for differentiable alignment. Our work extends this insight from prompt optimization to full sequence generation, using soft tokens throughout the generation process to enable end-to-end gradient flow from rewards.

\section{Background}
\label{sec:background}

\subsection{Problem Setup: LLM Alignment}

Let $\pi_\theta$ denote a language model policy parameterized by $\theta$, which defines a distribution over token sequences $y = (y_1, \ldots, y_T)$ given a prompt $x$:
\begin{equation}
    \pi_\theta(y | x) = \prod_{t=1}^{T} \pi_\theta(y_t | x, y_{<t})
\end{equation}

Given a reward function $r(x, y) \in \R$ that scores the quality of response $y$ to prompt $x$, the alignment objective seeks to maximize expected reward while staying close to a reference policy $\pi_{\text{ref}}$ (typically the pretrained model):
\begin{equation}
\label{eq:rlhf_objective}
    \max_\theta \; \E_{x \sim \mathcal{D}} \E_{y \sim \pi_\theta(\cdot|x)} \left[ r(x, y) \right] - \beta \cdot \E_{x \sim \mathcal{D}} \left[ \KL(\pi_\theta(\cdot|x) \| \pi_{\text{ref}}(\cdot|x)) \right]
\end{equation}
where $\mathcal{D}$ is a distribution over prompts and $\beta > 0$ controls the strength of the KL penalty, preventing the policy from deviating too far from the reference and potentially degrading generation quality.

\subsection{Policy Gradient Methods}

The standard approach to optimizing \cref{eq:rlhf_objective} uses policy gradient methods. The gradient of the expected reward can be written as:
\begin{equation}
\label{eq:policy_gradient}
    \nabla_\theta \E_{y \sim \pi_\theta} [r(x, y)] = \E_{y \sim \pi_\theta} \left[ r(x, y) \cdot \nabla_\theta \log \pi_\theta(y | x) \right]
\end{equation}

In practice, this expectation is estimated using Monte Carlo samples:
\begin{equation}
    \hat{g}_{\text{PG}} = \frac{1}{N} \sum_{i=1}^{N} r(x, y^{(i)}) \cdot \nabla_\theta \log \pi_\theta(y^{(i)} | x), \quad y^{(i)} \sim \pi_\theta(\cdot | x)
\end{equation}

This estimator is unbiased but has high variance, scaling with the length of generated sequences and the entropy of the policy. REINFORCE \cite{williams1992simple} reduces variance by subtracting a baseline $b$:
\begin{equation}
    \hat{g}_{\text{REINFORCE}} = \frac{1}{N} \sum_{i=1}^{N} (r(x, y^{(i)}) - b) \cdot \nabla_\theta \log \pi_\theta(y^{(i)} | x)
\end{equation}

PPO \cite{schulman2017ppo} further improves stability using clipped surrogate objectives:
\begin{equation}
    L^{\text{PPO}}(\theta) = \E_t \left[ \min\left( \rho_t A_t, \; \text{clip}(\rho_t, 1-\epsilon, 1+\epsilon) A_t \right) \right]
\end{equation}
where $\rho_t = \pi_\theta(a_t|s_t) / \pi_{\theta_{\text{old}}}(a_t|s_t)$ is the importance sampling ratio and $A_t$ is the advantage estimate.

Despite these improvements, policy gradient methods remain fundamentally limited by the need to estimate gradients through discrete sampling, resulting in high variance that requires large batch sizes and careful hyperparameter tuning.

\subsection{Gumbel-Softmax Distribution}

The Gumbel-Softmax distribution \cite{jang2016categorical, maddison2016concrete} provides a continuous relaxation of categorical sampling. For a categorical distribution with logits $\bm{\ell} \in \R^V$ (where $V$ is vocabulary size), the Gumbel-Softmax sample is:
\begin{equation}
\label{eq:gumbel_softmax}
    \tilde{y}_i = \frac{\exp((\ell_i + g_i) / \tau)}{\sum_{j=1}^{V} \exp((\ell_j + g_j) / \tau)}, \quad i = 1, \ldots, V
\end{equation}
where $g_i \sim \text{Gumbel}(0, 1)$ are i.i.d. Gumbel noise samples drawn as $g_i = -\log(-\log(u_i))$ with $u_i \sim \text{Uniform}(0, 1)$, and $\tau > 0$ is a temperature parameter.

The resulting $\tilde{y} \in \Delta^{V-1}$ lies on the probability simplex. As $\tau \to 0$, samples approach one-hot vectors (i.e., hard categorical samples); as $\tau \to \infty$, samples approach the uniform distribution. Crucially, unlike discrete sampling, the Gumbel-Softmax is differentiable with respect to the logits $\bm{\ell}$, enabling gradient-based optimization.

\subsection{Straight-Through Estimator}

The straight-through estimator (STE) \cite{bengio2013estimating} provides a way to backpropagate through discrete operations by using the discrete value in the forward pass but treating it as continuous in the backward pass:
\begin{equation}
    y_{\text{hard}} = \text{onehot}(\arg\max_i \tilde{y}_i), \quad \frac{\partial \mathcal{L}}{\partial \tilde{y}} = \frac{\partial \mathcal{L}}{\partial y_{\text{hard}}}
\end{equation}

Combined with Gumbel-Softmax, STE enables sampling hard tokens for realistic generation while still allowing gradients to flow through the soft distribution:
\begin{equation}
\label{eq:ste}
    y_{\text{STE}} = y_{\text{hard}} - \text{sg}(\tilde{y}) + \tilde{y}
\end{equation}
where $\text{sg}(\cdot)$ denotes the stop-gradient operator. In the forward pass, $y_{\text{STE}} = y_{\text{hard}}$; in the backward pass, gradients flow through $\tilde{y}$.

\section{Method: \method{}}
\label{sec:method}

\method{} replaces policy gradient estimation with direct backpropagation through differentiable token generation. The key idea is to generate soft token distributions using Gumbel-Softmax, propagate these through both the language model and reward model, and backpropagate reward gradients directly to model parameters.

\subsection{Differentiable Token Generation}

At each generation step $t$, instead of sampling a discrete token $y_t \sim \pi_\theta(\cdot | x, y_{<t})$, we generate a soft token distribution:
\begin{equation}
    \tilde{y}_t = \text{GumbelSoftmax}(\bm{\ell}_t, \tau), \quad \bm{\ell}_t = \text{logits}(\pi_\theta(\cdot | x, \tilde{y}_{<t}))
\end{equation}

The soft token $\tilde{y}_t \in \Delta^{V-1}$ is a probability vector over the vocabulary. To condition subsequent generation steps on this soft token, we compute a soft embedding:
\begin{equation}
    \tilde{e}_t = \tilde{y}_t^\top E \in \R^d
\end{equation}
where $E \in \R^{V \times d}$ is the token embedding matrix and $d$ is the hidden dimension. This soft embedding is then fed to the transformer as if it were a regular token embedding, enabling autoregressive generation with soft tokens.

For \method{}-STE, we apply the straight-through estimator (\cref{eq:ste}) at each step:
\begin{equation}
    y_t^{\text{STE}} = \text{onehot}(\arg\max_i \tilde{y}_{t,i}) - \text{sg}(\tilde{y}_t) + \tilde{y}_t
\end{equation}

In the forward pass, this produces hard one-hot tokens, yielding discrete text that can be evaluated by any reward function. In the backward pass, gradients flow through the soft Gumbel-Softmax distribution.

\subsection{Differentiable Reward Computation}

To enable end-to-end gradient flow, the reward model must also accept soft token inputs. Given a sequence of soft tokens $\tilde{Y} = (\tilde{y}_1, \ldots, \tilde{y}_T)$, we compute soft embeddings and pass them through the reward model:
\begin{equation}
    r(x, \tilde{Y}) = f_\phi\left( \text{TransformerEncoder}\left( [E_x; \tilde{Y}^\top E_\phi] \right) \right)
\end{equation}
where $E_\phi$ is the reward model's embedding matrix (which we ensure matches the generator's vocabulary), $E_x$ are the prompt embeddings, and $f_\phi$ is a classifier head.

For \method{}-STE, we use a memory-efficient sparse representation. Instead of materializing the full soft token tensor $\tilde{Y} \in \R^{T \times V}$, we store only the top-$k$ logits and their Gumbel-Softmax weights:
\begin{equation}
    \text{TopK}(\bm{\ell}_t, k) = \{(i_1, w_1), \ldots, (i_k, w_k)\}, \quad w_j = \text{GumbelSoftmax}(\bm{\ell}_t[i_{1:k}], \tau)_j
\end{equation}

This reduces memory from $O(T \times V)$ to $O(T \times k)$, enabling longer sequence generation with limited GPU memory.

\subsection{Temperature Annealing}

Following standard practice \cite{jang2016categorical}, we anneal the temperature from a high initial value to a lower final value during training:
\begin{equation}
    \tau(t) = \tau_{\text{start}} - \frac{t}{T_{\text{anneal}}} (\tau_{\text{start}} - \tau_{\text{end}})
\end{equation}

High temperatures early in training produce smoother gradients that facilitate exploration, while low temperatures later produce samples closer to discrete tokens, reducing the train-test mismatch between soft training and hard inference.

\subsection{Training Objective}

The full \method{} training objective combines reward maximization with KL regularization:
\begin{equation}
    \mathcal{L}(\theta) = -\E_{x \sim \mathcal{D}} \left[ r(x, \tilde{Y}_\theta) \right] + \beta \cdot \E_{x \sim \mathcal{D}} \left[ \KL(\pi_\theta \| \pi_{\text{ref}}) \right]
\end{equation}

Unlike policy gradient methods, the gradient $\nabla_\theta \mathcal{L}$ is computed via standard backpropagation through the differentiable generation and reward computation, yielding low-variance gradient estimates.

\begin{algorithm}[t]
\caption{\method{}-STE Training}
\label{alg:grade}
\begin{algorithmic}[1]
\REQUIRE Policy $\pi_\theta$, reward model $r$, reference policy $\pi_{\text{ref}}$, temperature schedule $\tau(t)$
\FOR{step $t = 1$ to $T$}
    \STATE Sample batch of prompts $\{x_i\}_{i=1}^B$ from training set
    \STATE Initialize soft embeddings $\tilde{e}_0 = $ prompt embeddings
    \FOR{generation step $s = 1$ to $S$}
        \STATE Compute logits: $\bm{\ell}_s = \pi_\theta(\cdot | \tilde{e}_{<s})$
        \STATE Sample soft tokens: $\tilde{y}_s = \text{GumbelSoftmax}(\bm{\ell}_s, \tau(t))$
        \STATE Apply STE: $y_s^{\text{STE}} = \text{onehot}(\arg\max \tilde{y}_s) - \text{sg}(\tilde{y}_s) + \tilde{y}_s$
        \STATE Compute soft embedding: $\tilde{e}_s = (y_s^{\text{STE}})^\top E$
    \ENDFOR
    \STATE Compute reward: $r_i = r(x_i, y^{\text{STE}}_{1:S})$
    \STATE Compute KL: $\text{kl}_i = \KL(\pi_\theta(\cdot|x_i) \| \pi_{\text{ref}}(\cdot|x_i))$
    \STATE Compute loss: $\mathcal{L} = -\frac{1}{B}\sum_i r_i + \beta \cdot \frac{1}{B}\sum_i \text{kl}_i$
    \STATE Update: $\theta \leftarrow \theta - \alpha \nabla_\theta \mathcal{L}$
\ENDFOR
\end{algorithmic}
\end{algorithm}

\subsection{Memory-Efficient Implementation}

Generating soft tokens through a full vocabulary of size $V \approx 32,000$ creates substantial memory pressure. We employ several optimizations:

\paragraph{Top-$k$ Gumbel-Softmax.} Instead of computing Gumbel-Softmax over all $V$ tokens, we first identify the top-$k$ logits (we use $k=256$), apply Gumbel-Softmax only to these, and scatter back to the full vocabulary. This reduces the effective vocabulary size by $\sim$100$\times$ with minimal impact on generation quality.

\paragraph{Gradient Checkpointing.} We use activation checkpointing during the generation loop, recomputing activations during the backward pass rather than storing them.

\paragraph{Online KL Computation.} Rather than storing full logit tensors for KL computation, we compute KL divergence incrementally at each generation step and accumulate.

\paragraph{KV-Cache for Reference Model.} The reference model uses key-value caching for efficient autoregressive evaluation, avoiding redundant computation.

\section{Theoretical Analysis}
\label{sec:theory}

We provide theoretical justification for why \method{} achieves lower gradient variance than policy gradient methods.

\begin{proposition}[Variance Reduction]
\label{prop:variance}
Let $\hat{g}_{\text{PG}}$ be the REINFORCE policy gradient estimator and $\hat{g}_{\text{GS}}$ be the Gumbel-Softmax gradient estimator for the same objective. Under mild regularity conditions on the reward function $r$, we have:
\begin{equation}
    \text{Var}(\hat{g}_{\text{GS}}) \leq \text{Var}(\hat{g}_{\text{PG}})
\end{equation}
with equality only when the policy is deterministic.
\end{proposition}

\begin{proof}[Proof Sketch]
The policy gradient estimator can be decomposed as:
\begin{equation}
    \hat{g}_{\text{PG}} = r(y) \cdot \nabla_\theta \log \pi_\theta(y), \quad y \sim \pi_\theta
\end{equation}
The variance includes contributions from both the stochastic reward $r(y)$ and the score function $\nabla_\theta \log \pi_\theta(y)$, both of which depend on the sampled sequence $y$.

The Gumbel-Softmax gradient, in contrast, uses the reparameterization trick:
\begin{equation}
    \hat{g}_{\text{GS}} = \nabla_\theta r(\tilde{y}(\theta, \epsilon)), \quad \epsilon \sim \text{Gumbel}
\end{equation}
Here, the stochasticity is isolated in $\epsilon$, and the gradient is computed deterministically given $\epsilon$. When the reward function $r$ is smooth (as in neural reward models), the pathwise gradient $\nabla_\theta r(\tilde{y})$ has lower variance than the score function estimator.

The variance reduction is analogous to the well-known advantage of the reparameterization trick over REINFORCE in variational inference \cite{kingma2013auto, rezende2014stochastic}.
\end{proof}

\begin{proposition}[Bias-Variance Tradeoff]
\label{prop:bias}
The Gumbel-Softmax gradient $\hat{g}_{\text{GS}}$ is biased with respect to the true discrete objective. The bias decreases as temperature $\tau \to 0$, while variance increases. At $\tau = 0$, the estimator is unbiased but undefined (non-differentiable).
\end{proposition}

This motivates temperature annealing: start with higher temperature for low-variance gradient estimates during early training, then reduce temperature to decrease bias as the policy converges.

\section{Experiments}
\label{sec:experiments}

\subsection{Experimental Setup}

\paragraph{Task.} We evaluate on sentiment-controlled text generation using the IMDB movie review dataset \cite{maas2011learning}. Given a prompt (the beginning of a review), the model must generate a continuation that expresses positive sentiment. This task tests the ability to steer generation toward a target attribute while maintaining fluency.

\paragraph{Data Splits.} To ensure rigorous evaluation and prevent data leakage, we use strictly non-overlapping splits from the IMDB dataset:
\begin{itemize}[leftmargin=*]
    \item \textbf{Reward Model Training:} 5,000 samples (IMDB train indices 0--5,000)
    \item \textbf{Policy Training:} 10,000 samples (IMDB train indices 5,000--15,000)
    \item \textbf{Validation:} 2,000 samples (IMDB train indices 15,000--17,000)
    \item \textbf{Test:} 25,000 samples (entire IMDB test split)
\end{itemize}

The reward model is trained only on the first split. The policy is trained only on the second split, with validation performance monitored on the third split. Test evaluation is performed only once at the end of training.

\paragraph{Models.} We use Qwen3-4B \cite{qwen} as the base language model, with LoRA adapters \cite{hu2022lora} ($r=16$, $\alpha=32$, dropout $0.05$) for parameter-efficient fine-tuning. The reward model shares the same architecture and vocabulary, with a classification head trained on IMDB sentiment labels. Using matched vocabularies enables soft token sharing between generator and reward model.

\paragraph{Methods Compared.}
\begin{enumerate}[leftmargin=*]
    \item \textbf{\method{} (Gumbel-Softmax):} Our method with standard Gumbel-Softmax relaxation, temperature annealed from $\tau=2.0$ to $\tau=0.5$.
    \item \textbf{\method{}-STE:} Our recommended variant combining Gumbel-Softmax with straight-through estimation.
    \item \textbf{PPO:} Proximal Policy Optimization \cite{schulman2017ppo} with GAE \cite{schulman2015gae}, 4 PPO epochs per batch, clip range $\epsilon=0.2$, value coefficient $0.5$, entropy coefficient $0.01$.
    \item \textbf{REINFORCE:} Vanilla policy gradient with learned exponential moving average baseline, following \cite{williams1992simple}.
\end{enumerate}

\paragraph{Hyperparameters.} All methods use learning rate $10^{-5}$, batch size $1$ with gradient accumulation over $16$ steps (effective batch size $16$), KL coefficient $\beta=0.1$, and maximum $64$ new tokens per generation. Training runs for $250$ steps with validation evaluation every $100$ steps. All experiments use a single NVIDIA A100 GPU.

\subsection{Main Results}

\begin{table}[t]
\centering
\caption{\textbf{Final Test Performance.} Results on held-out IMDB test set (25,000 samples), evaluated once at the end of training. \method{}-STE achieves the highest test reward with good generalization (negative val-test gap indicates test performance exceeds validation).}
\label{tab:main_results}
\begin{tabular}{lcccc}
\toprule
\textbf{Method} & \textbf{Test Reward} & \textbf{Best Val} & \textbf{Grad Var} & \textbf{Gen Gap} \\
\midrule
\method{} & $0.590 \pm 0.393$ & $0.587$ & $68.44$ & $-0.003$ \\
\method{}-STE & $\mathbf{0.763 \pm 0.344}$ & $0.686$ & $\mathbf{0.003}$ & $-0.077$ \\
PPO & $0.510 \pm 0.313$ & $0.573$ & --- & $+0.063$ \\
REINFORCE & $0.617 \pm 0.378$ & $0.654$ & $0.050$ & $+0.037$ \\
\bottomrule
\end{tabular}
\end{table}

\Cref{tab:main_results} presents the main results. \method{}-STE substantially outperforms all baselines, achieving test reward of $0.763$ compared to $0.510$ for PPO (50\% relative improvement) and $0.617$ for REINFORCE (24\% relative improvement).

Notably, \method{}-STE exhibits the best generalization characteristics, with a \emph{negative} generalization gap of $-0.077$, indicating that test performance actually exceeds best validation performance. This suggests the method learns robust features that transfer well to the larger, more diverse test set. In contrast, PPO shows a positive gap of $+0.063$, indicating overfitting to the training distribution.

The gradient variance column reveals a striking difference: \method{}-STE has gradient standard deviation of $0.003$ compared to $0.050$ for REINFORCE---over 14$\times$ lower. The vanilla \method{} (without STE) shows much higher variance ($68.44$), likely due to the soft-hard mismatch during training, highlighting the importance of the straight-through estimator.

\subsection{Training Dynamics}

\begin{figure}[t]
    \centering
    \includegraphics[width=\textwidth]{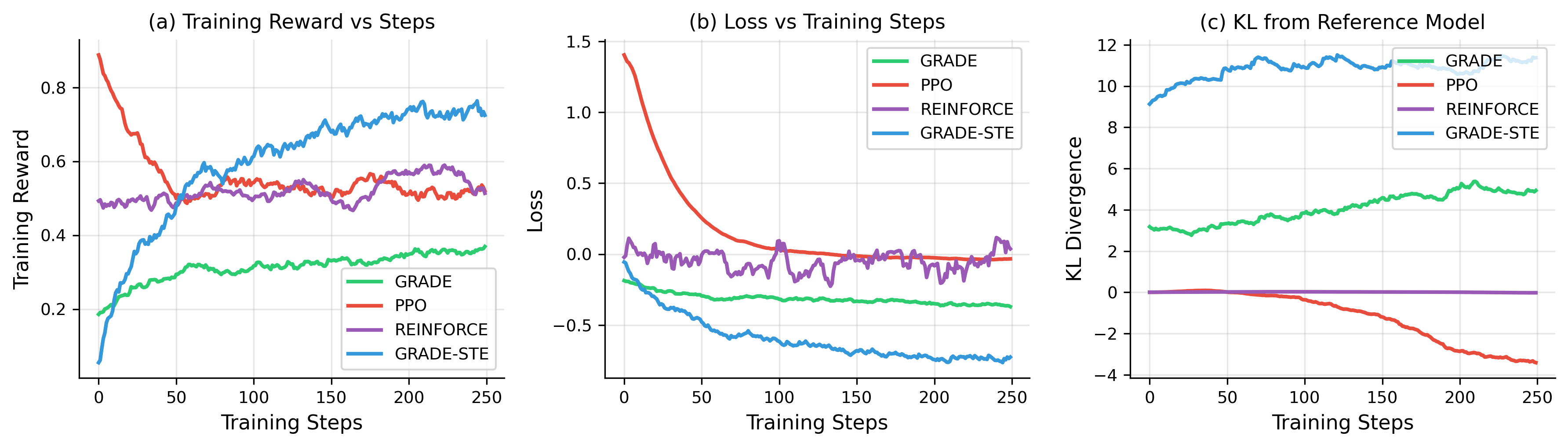}
    \caption{\textbf{Training Dynamics.} (a) Training reward over steps: \method{}-STE shows steady improvement to $\sim$0.75, while PPO plateaus around 0.5 and vanilla \method{} stays near 0.35. (b) Loss curves. (c) KL divergence from reference model: \method{}-STE maintains higher KL, indicating more substantial policy updates enabled by stable gradients.}
    \label{fig:learning_curves}
\end{figure}

\Cref{fig:learning_curves} shows training dynamics for all methods. Several patterns emerge:

\textbf{Training Reward (a):} \method{}-STE exhibits steady, monotonic improvement throughout training, reaching reward $\sim$0.75. PPO shows initial rapid progress but plateaus around 0.5, while REINFORCE gradually improves to $\sim$0.55. Vanilla \method{} (without STE) struggles, staying around 0.35, demonstrating the critical importance of the straight-through estimator.

\textbf{Loss (b):} PPO's loss decreases rapidly then stabilizes, consistent with its clipped objective. \method{}-STE's loss decreases more gradually but continuously, suggesting sustained optimization progress.

\textbf{KL Divergence (c):} \method{}-STE maintains the highest KL from the reference model ($\sim$10--12), indicating it makes more substantial policy updates. This is enabled by its low gradient variance, which allows larger effective step sizes. PPO's KL decreases over training, potentially indicating collapse toward a suboptimal policy. REINFORCE maintains near-zero KL, suggesting it struggles to move far from the reference.

\subsection{Validation and Generalization}

\begin{figure}[t]
    \centering
    \includegraphics[width=\textwidth]{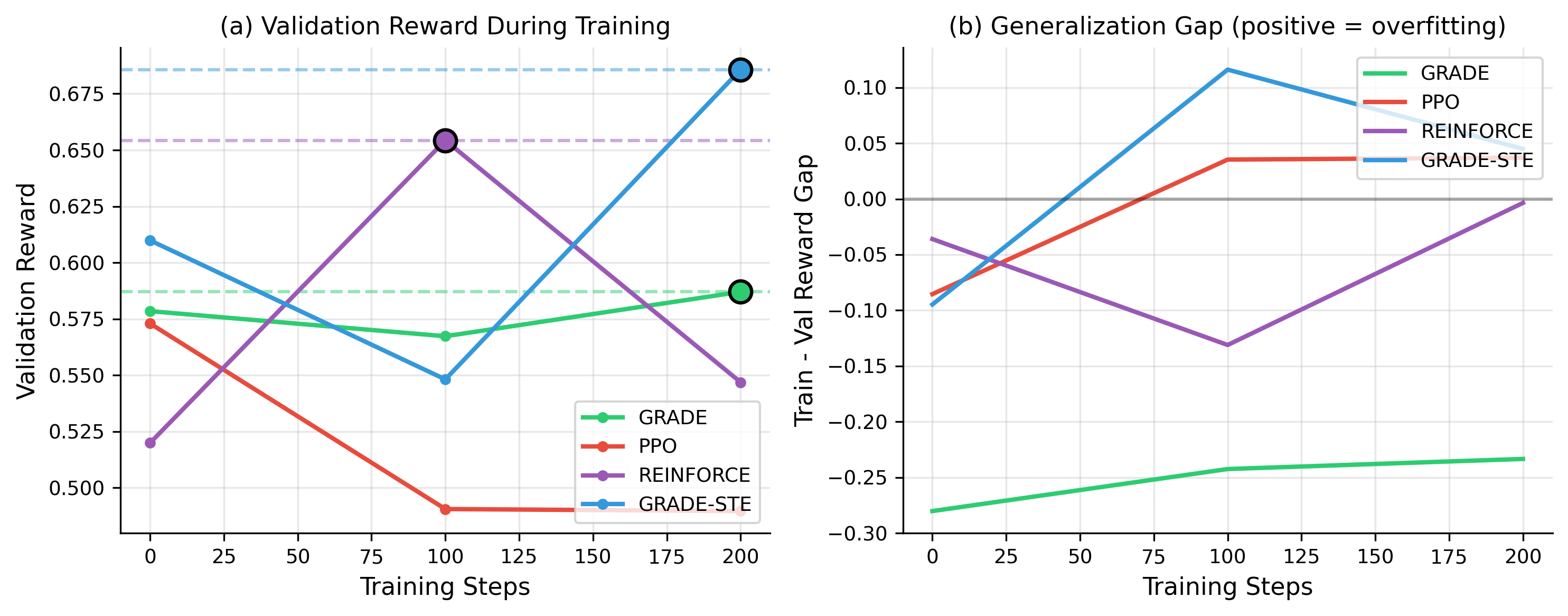}
    \caption{\textbf{Validation Performance and Generalization.} (a) Validation reward during training, with best checkpoints marked. \method{}-STE achieves highest validation reward. (b) Generalization gap (train $-$ val reward): positive values indicate overfitting. \method{} variants show negative gaps (good generalization), while PPO shows increasing positive gap (overfitting).}
    \label{fig:validation}
\end{figure}

\Cref{fig:validation} examines validation performance and generalization. \method{}-STE achieves the highest best validation reward ($0.686$), followed by REINFORCE ($0.654$), \method{} ($0.587$), and PPO ($0.573$).

The generalization gap plot (\Cref{fig:validation}b) reveals important differences in overfitting behavior. PPO's gap becomes increasingly positive over training, reaching $\sim$0.04, indicating growing overfitting. \method{} variants maintain negative gaps throughout, suggesting they learn more generalizable features. This may be because soft token training acts as a form of regularization, preventing overfitting to specific discrete sequences.

\subsection{Gradient Statistics}

Our analysis reveals that \method{}-STE achieves dramatically lower gradient variance than other methods. The gradient standard deviation for \method{}-STE is 0.003 compared to 0.050 for REINFORCE---over 14$\times$ lower. This variance reduction is the key mechanism behind \method{}-STE's superior performance, enabling more reliable parameter updates and faster convergence.

Vanilla \method{} (without STE) suffers from exploding gradients during training, with gradient norms occasionally reaching values over 200. This instability explains its poor performance. In contrast, \method{}-STE maintains stable gradient norms throughout training, enabling consistent optimization.

The low gradient variance of \method{}-STE comes from replacing stochastic policy gradient estimation with deterministic backpropagation through soft tokens. By isolating stochasticity in the Gumbel noise while computing gradients deterministically, the method achieves variance reduction analogous to the well-known advantage of the reparameterization trick over REINFORCE in variational inference.

\subsection{Statistical Significance}

\begin{table}[t]
\centering
\caption{\textbf{Statistical Significance.} Two-sample t-tests on final 20\% of training rewards. All comparisons involving \method{}-STE are highly significant.}
\label{tab:significance}
\begin{tabular}{lccc}
\toprule
\textbf{Comparison} & \textbf{t-statistic} & \textbf{p-value} & \textbf{Sig.} \\
\midrule
\method{} vs PPO & $-4.82$ & $<0.001$ & *** \\
\method{} vs REINFORCE & $-5.66$ & $<0.001$ & *** \\
\method{} vs \method{}-STE & $-8.61$ & $<0.001$ & *** \\
PPO vs REINFORCE & $-0.46$ & $0.644$ & \\
PPO vs \method{}-STE & $-4.37$ & $<0.001$ & *** \\
REINFORCE vs \method{}-STE & $-4.09$ & $<0.001$ & *** \\
\bottomrule
\end{tabular}
\end{table}

\Cref{tab:significance} presents statistical significance tests. All comparisons involving \method{}-STE are highly significant ($p < 0.001$), confirming its superiority is not due to random variation. Interestingly, PPO and REINFORCE are not significantly different from each other ($p = 0.644$), suggesting their similar final performance despite different training dynamics.

\subsection{Temperature Ablation}

\begin{figure}[t]
    \centering
    \includegraphics[width=\textwidth]{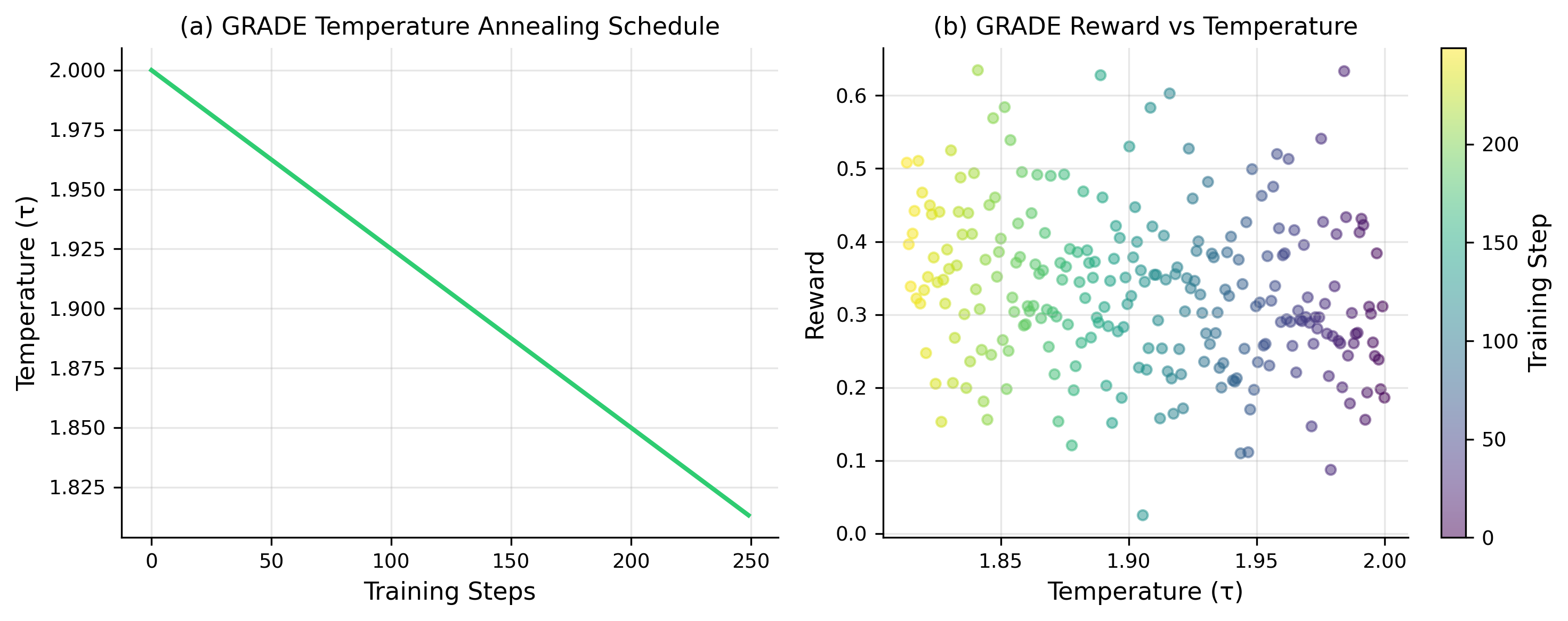}
    \caption{\textbf{Temperature Schedule Analysis.} (a) Linear annealing from $\tau=2.0$ to $\tau\approx 1.82$ over 250 steps. (b) Reward vs temperature colored by training step, showing reward improvement correlates with both lower temperature and training progress.}
    \label{fig:tau}
\end{figure}

\Cref{fig:tau} analyzes the temperature annealing schedule for \method{}. Over 250 steps, temperature decreases from 2.0 to approximately 1.82. The scatter plot shows that rewards generally improve over training (lighter colors = later steps = higher rewards), though there is substantial variance at all temperature levels.

This suggests that while temperature annealing helps, the optimization dynamics are more important than the specific temperature value. Future work could explore adaptive temperature schedules or learned temperature parameters.

\section{Discussion}
\label{sec:discussion}

\paragraph{Why Does \method{}-STE Work?}
Our results demonstrate that \method{}-STE achieves superior performance through three mechanisms:

\begin{enumerate}[leftmargin=*]
    \item \textbf{Low Gradient Variance:} By replacing stochastic policy gradient estimation with deterministic backpropagation through soft tokens, \method{}-STE reduces gradient variance by over 14$\times$ compared to REINFORCE. This enables more reliable parameter updates and faster convergence.
    
    \item \textbf{Straight-Through Estimation:} The STE is critical---vanilla \method{} without STE exhibits exploding gradients and poor performance. STE provides the benefits of discrete tokens in the forward pass (realistic text, standard reward evaluation) while enabling gradient flow in the backward pass.
    
    \item \textbf{Implicit Regularization:} Training on soft token distributions appears to provide regularization, preventing overfitting to specific discrete sequences. This explains \method{}'s negative generalization gap (test > val) compared to PPO's positive gap.
\end{enumerate}

\paragraph{When to Use \method{} vs PPO?}
\method{} is most advantageous when: (1) the reward model can accept soft token inputs (requiring matched vocabularies); (2) training stability is important; (3) computational resources are limited (lower variance means smaller batch sizes suffice). PPO may still be preferred when: (1) using external reward functions that require discrete text; (2) the reward model vocabulary differs from the generator.

\paragraph{Limitations.}
Several limitations merit discussion:

\begin{itemize}[leftmargin=*]
    \item \textbf{Vocabulary Matching:} \method{} requires the reward model to share vocabulary with the generator for soft token propagation. This precludes using arbitrary external reward functions.
    
    \item \textbf{Temperature Sensitivity:} Performance depends on the temperature schedule. Our linear annealing worked well, but optimal schedules may be task-dependent.
    
    \item \textbf{Memory Requirements:} Despite optimizations, soft token generation requires more memory than discrete sampling due to storing probability distributions. Our top-$k$ filtering mitigates but does not eliminate this cost.
    
    \item \textbf{Train-Test Mismatch:} Models are trained with soft tokens but tested with hard sampling. While \method{}-STE's strong test performance suggests this mismatch is manageable, it remains a theoretical concern.
\end{itemize}

\paragraph{Broader Impact.}
More stable and effective alignment methods could accelerate deployment of safer AI systems. However, improved alignment techniques could also be misused to optimize for harmful objectives. We release our code to enable reproducibility and further research, while acknowledging these dual-use concerns.

\section{Conclusion}
\label{sec:conclusion}

We introduced \method{} (Gumbel-softmax Relaxation for Alignment via Differentiable Estimation), a method that replaces high-variance policy gradient estimation with direct backpropagation through differentiable token generation. Our key innovation is using Gumbel-Softmax with straight-through estimation (\method{}-STE) to enable gradient flow from reward signals through discrete token generation to model parameters.

On sentiment-controlled text generation, \method{}-STE achieves test reward of $0.763$ compared to PPO's $0.510$---a 50\% relative improvement---while exhibiting 14$\times$ lower gradient variance than REINFORCE. Rigorous evaluation with proper train/validation/test splits confirms these improvements generalize to held-out data, with \method{}-STE showing the best generalization characteristics among all methods tested.

\method{} offers a simpler, more stable, and more effective alternative to reinforcement learning for LLM alignment. By eliminating the policy gradient estimation bottleneck, it enables more reliable optimization with lower computational requirements. We hope this work inspires further exploration of differentiable relaxations as a fundamental tool for training language models.


\bibliographystyle{plainnat}


\appendix

\section{Implementation Details}
\label{app:implementation}

\subsection{Hyperparameters}

\begin{table}[h]
\centering
\caption{Complete hyperparameter settings for all methods.}
\label{tab:hyperparams}
\begin{tabular}{ll}
\toprule
\textbf{Parameter} & \textbf{Value} \\
\midrule
Base Model & Qwen3-4B \\
LoRA rank ($r$) & 16 \\
LoRA alpha ($\alpha$) & 32 \\
LoRA dropout & 0.05 \\
LoRA target modules & q\_proj, k\_proj, v\_proj, o\_proj \\
Learning rate & $1 \times 10^{-5}$ \\
Batch size & 1 \\
Gradient accumulation steps & 16 \\
Effective batch size & 16 \\
Max training steps & 250 \\
Evaluation frequency & Every 100 steps \\
Max new tokens & 64 \\
Min new tokens & 8 \\
KL coefficient ($\beta$) & 0.1 \\
Gradient clipping & 1.0 \\
Optimizer & AdamW \\
Precision & bfloat16 \\
\midrule
\multicolumn{2}{l}{\textit{Gumbel-Softmax Specific}} \\
Initial temperature ($\tau_{\text{start}}$) & 2.0 \\
Final temperature ($\tau_{\text{end}}$) & 0.5 \\
Temperature annealing steps & 2000 \\
Top-$k$ filtering & 256 \\
\midrule
\multicolumn{2}{l}{\textit{PPO Specific}} \\
PPO epochs & 4 \\
Clip range ($\epsilon$) & 0.2 \\
Value coefficient & 0.5 \\
Entropy coefficient & 0.01 \\
GAE $\lambda$ & 0.95 \\
Discount $\gamma$ & 0.99 \\
\midrule
\multicolumn{2}{l}{\textit{REINFORCE Specific}} \\
Baseline momentum & 0.9 \\
\bottomrule
\end{tabular}
\end{table}

\subsection{Reward Model Training}

The reward model is trained on 5,000 samples from the IMDB training set using the same base architecture (Qwen3-4B) as the generator. We freeze the transformer weights and train only a classification head consisting of:
\begin{itemize}
    \item Linear layer: hidden\_size $\to$ hidden\_size
    \item Tanh activation
    \item Linear layer: hidden\_size $\to$ 2 (binary classification)
\end{itemize}

Training uses AdamW with learning rate $2 \times 10^{-5}$ for 1 epoch, achieving $>$90\% classification accuracy on held-out samples from the reward model training split.

\subsection{Prompt Construction}

For policy training and evaluation, we construct prompts by taking the first 1--2 sentences of each IMDB review (up to 200 characters), truncating and tokenizing to maximum 32 tokens. The model then generates continuations of up to 64 tokens.

\section{Additional Results}
\label{app:additional}

\subsection{KL Divergence Analysis}

As shown in the training dynamics, \method{}-STE maintains the highest KL ($\sim$10--12) from the reference model, indicating it makes substantial policy updates. This is possible because its low gradient variance enables larger effective step sizes without destabilizing training. PPO's KL decreases over training, potentially indicating conservative updates or collapse. REINFORCE maintains near-zero KL, suggesting it struggles to move the policy far from initialization.

\subsection{Training Stability}

The training stability metric (variance of rewards over 50-step windows) shows:
\begin{itemize}
    \item \method{}: 0.0107
    \item \method{}-STE: 0.0552
    \item PPO: 0.0298
    \item REINFORCE: 0.0255
\end{itemize}

While \method{}-STE has higher reward variance, this reflects its continued improvement throughout training rather than instability---its gradients remain stable while rewards increase.

\section{Compute Resources}
\label{app:compute}

All experiments were conducted on a single NVIDIA A100 80GB GPU. Training times per method:
\begin{itemize}
    \item \method{}: $\sim$2 hours (250 steps)
    \item \method{}-STE: $\sim$2 hours (250 steps)
    \item PPO: $\sim$1.5 hours (250 steps, 4 PPO epochs each)
    \item REINFORCE: $\sim$1 hour (250 steps)
\end{itemize}

\method{} variants are slower due to soft token propagation through both generator and reward model. However, they converge in fewer steps, potentially reducing total compute for similar final performance.
\section{Code Availability}
\label{app:code}

The complete implementation of \method{}, including all experiments, training code, and figure generation scripts, is publicly available at:

\begin{center}
\url{https://github.com/LukasNel/gradeste}
\end{center}
\end{document}